# Semantic Detection of Potential Wind-borne Debris in Construction Jobsites: Digital Twining for Hurricane Preparedness and Jobsite Safety

Mirsalar Kamari[1] and Youngjib Ham, Ph.D. A.M.ASCE[2]

[1] PhD Student, Department of Construction Science, Texas A&M University, 3137 TAMU, College Station, TX 77843; email: kamari@tamu.edu

[2] Assistant Professor, Department of Construction Science, Texas A&M University, Francis Hall 329B, 3137 TAMU, College Station, TX 77843; email: yham@tamu.edu

**ABSTRACT**

In the United States, hurricanes are the most devastating natural disasters causing billions of dollars worth of damage every year. More importantly, construction jobsites are classified among the most vulnerable environments to severe wind events. During hurricanes, unsecured and incomplete elements of construction sites, such as scaffoldings, plywoods, and metal rods, will become the potential wind-borne debris, causing cascading damages to the construction projects and the neighboring communities. Thus, it is no wonder that construction firms implement jobsite emergency plans to enforce preparedness responses before extreme weather events. However, relying on checklist-based emergency action plans to carry out a thorough hurricane preparedness is challenging in large-scale and complex site environments. For enabling systematic responses for hurricane preparedness, we have proposed a vision-based technique to identify and analyze the potential wind-borne debris in construction jobsites. Building on this, this paper demonstrates the fidelity of a new machine vision-based method to support construction site hurricane preparedness and further discuss its implications. The outcomes indicate that the convenience of visual data collection and the advantages of the machine vision-based frameworks enable rapid scene understanding and thus, provide critical heads up for practitioners to recognize and localize the potential wind-borne derbies in construction jobsites and effectively implement hurricane preparedness.

**INTRODUCTION**

Construction is widely regarded as one of the most dangerous occupations, as the Occupational Health and Safety Administration reported that 1 in 5 of total worker fatalities recorded in 2019 was in construction (OSHA 2019). Among those documented accidents in the United States, wind-induced accidents and damages account for two-third of insurance losses in the construction industry (CLM 2018). As prior studies suggested, construction sites are primary victims of hurricanes (Kamari and Ham 2020). For example, in 2012, as Hurricane Sandy ravaged downtown New York City, the construction of the world trade center suffered 185 million dollars of damage (Fermino 2013). Moreover, category five Hurricane Wilma wrecked southern Florida and impacted the construction of the Miami International Airport project, leaving behind about 4.5 million dollars of damage (FEMA 2012). While such damages are typically accounted for flooding and debris strikes, (Minor 2005) indicated that the construction debris damage is an important part of the total loss during extreme wind events. In order to minimize the potential risk of construction debris during a hurricane event, cities in the state of Florida such as Bradenton, Venice, and West Palm Beach have enacted ordinances that prohibit contractors from leaving unprotected and loose materials in their jobsites once a hurricane watch warning is issued. Moreover, violators, who fail to remove and relocate the potential wind-borne debris, are classified as second-degree misdemeanors and are subjected to fines and imprisonment of up to 60 days (Bohm 2020). Preparing jobsites for extreme wind events in a quick and accurate fashion has accordingly become an ongoing objective for the construction industry, with the ultimate goal of reducing the damage to construction project sites as well as neighboring communities. In the wake of recent growing concerns regarding hurricane preparedness, in this research, building on digital twinning and volumetric measurement techniques, we propose a novel vision-based method to detect and analyze the potential wind-borne debris in construction jobsites. The main contribution of this research is encoding the context of potential wind-borne debris into machine vision algorithms to automatically flag and evaluate entailed risk of debris in complex and ever-changing construction jobsites. The outcome of the proposed method is a heatmap to associate the risk of potential wind-borne debris with respect to the severity of the wind event. The proposed rapid scene understanding technique will provide crucial heads up for practitioners to quickly locate the risk of potential wind-borne debris prior to hurricanes and thereby implementing proactive measures in a timely manner.

**BACKGROUND**

A great body of literature on disaster management have addressed post-disaster efforts , including damage assessment (Ghorbani and Behzadan 2020), recovery plans, while pre-disaster measures are not widely explored yet (Kamari and Ham 2020; Sherafat and Rashidi 2019). Unlike post-disaster measures, the main objective of pre-disaster efforts is to understand and estimate the severity of potential risk, and thereby taking proactive measures to mitigate the potential damages. In this sense, (Gregg et al. 2004) classified the risk of hazard-prone environments based on their geographical location and historical records. In the context of hurricane damage estimation, (Letchford et al. 2000) presented a case study and investigated the debris-induced damages after Fort Worth Tornado, and concluded that debris size and density are among the main grounds to estimate the damage during the wind events. Later, in accordance with the observation of (Letchford et al. 2000), a ridable framework to forecast the risk of debris was proposed by (Wills et al. 2002), and they classified the risk of debris with respect to their characteristics such as shape and volume. Building on such insights, construction companies, safety directors, and contractors develop hurricane preparedness checklists to identify the risk associated with potential wind-borne debris before hurricanes. However, given that there is a limited time to carry out hurricane preparedness, developing a systematic approach to automatically uncover the risk of debris in construction sites is not a trivial task. To enable such a systematic approach, a growing body of literature presents the recognition of risk through collected images (Sherafat et al. 2020; Sherafat and Rashidi 2020). However, the outcome of prior vision-based approaches that enable detection of debris within images is not expected to translate into desirable outcomes in the context of hurricane preparedness, as information regarding debris volume and their location is still needed for the risk assessment. Given that camera-equipped platforms, such as Unmanned Aerial Vehicles (UAVs), are commonly used in the construction industry (Ham and Kamari 2019; Kamari and

Gunes 2016; Kamari and Ham 2018), we leverage visual data to retrieve the attributes of debris, and enable systematic and risk-informed approach for hurricane preparedness.

**METHODOLOGY**

The proposed framework is built upon the following modules: 1) semantic digital twinning of construction sites, and 2) risk assessment of debris based on their volumes, as illustrated in Figure 1. In the following sections, each module is introduced, and a case study is presented to evaluate the proposed model.

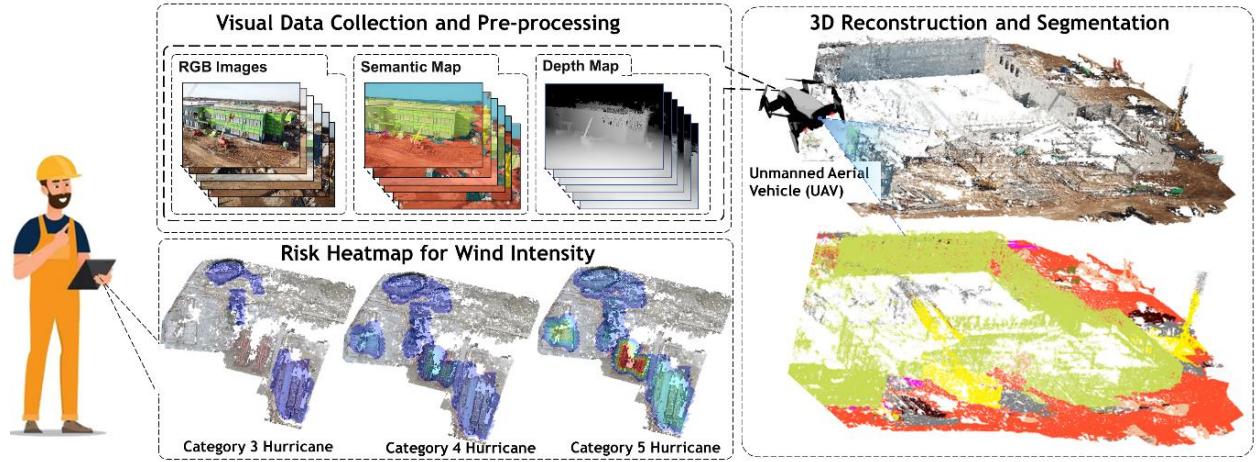

**Figure 1. Overview of the proposed method.**

Digital twinging is the process of generating a 3D virtual representation of an asset or object (Kim and Ham 2020; Kim et al. 2021). In order to represent the at-risk jobsite environment, we build upon the Structure from Motion (SfM) technique with Scale-Invariant Feature Transformation (SIFT) (Lowe 2004). The technique investigates pairwise matching among images to compute fundamental matrix and retrieves extrinsic characteristics, including translation and rotation of the camera. The location of points representing successful feature matches among images is then computed in the 3D. In order to populate the density of the point cloud model, images are divided into patches, and the iterative framework based on match, expand, and filter is used to refine the density of the point cloud model (Farhadmanesh et al. 2021; Kamari and Ham 2020). In order to enforce the detection of potential wind-borne debris in 3D, we detect the potential wind-borne debris in 2D images and project the 2D semantic outcomes onto the 3D point cloud. We build on the Resnet 18 deep learning model, containing 11 million parameters and 18 layers to achieve the pixel-level segmentation. Projection of such semantic values from 2D level to 3D point cloud is performed by establishing correspondence between coordinates of pixels within image and location of points within the point cloud model, as described in Equation (1).

$$C_i = K_{3\times3}[R_{3\times3}|T_{3\times1}]C_w \tag{1}$$

where, $C_i$ corresponds to the location of $i$th pixel in the segmented image, and $C_w$ is the location of its associated point in the point cloud model. The parameters $R, T$ represent the extrinsic attribute of the camera, namely the rotation, translation, and $K$ denotes the intrinsic characteristic of the camera. Figure 2 exemplifies the semantic projection onto the point cloud model. Camera positions retrieved during the reconstruction of the point cloud model are shown in blue, and the projection camera is shown in red (Figure 2.a). As observed in Figure 2.b, Equation (1) cannot handle shadows of projection, which leads to misclassifications of occluded objects. In order to account for occlusion during the semantic projection, we build upon extrinsic camera parameters and point cloud models as priors to compute depth maps. In this regard, a grid of patches is considered at the camera position. Through Equation (1), points visible at each patch are studied, and the closest point to the associating patch is retrieved. By performing such process for all patches, a depth map corresponding to the camera position is obtained. Figure 2.c demonstrates the depth-aware semantic projection. By performing the depth-aware semantic projection among all camera positions, a segmented digital twin model of potential wind-borne debris is obtained (Figure 2.d).

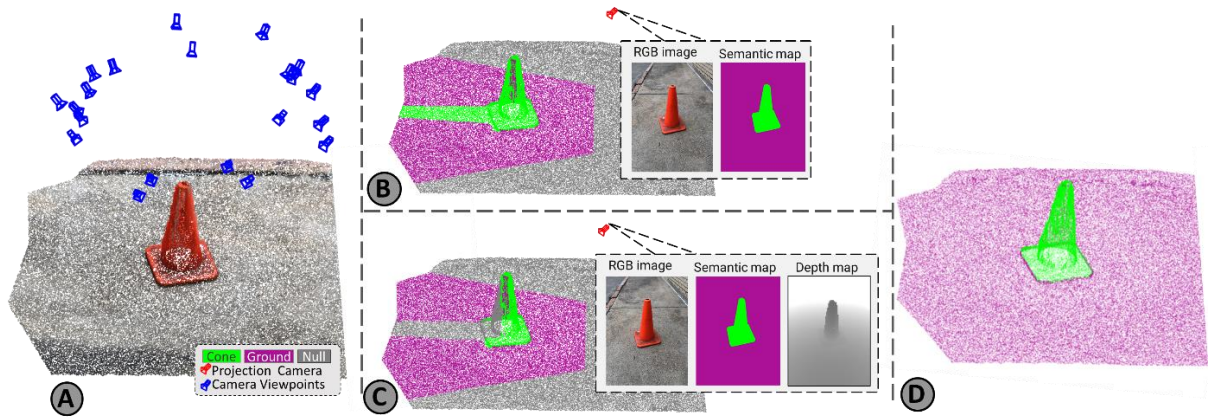

**Figure 2. Proposed semantic digital twining to detect objects of interest.**

The segmented digital twin model retains information regarding the potential wind-borne debris in construction sites, including their type and location. Upon successful semantic segmentation at the 3D level, we compute the volume of debris to visualize their associated risk map. Voluminous debris poses a greater threat during the wind event, and thus the size estimation of debris is a key step for the risk assessment. Prior works on point cloud processing suggested that the volume of a target object could be calculated reliably in a segmented point cloud model. Building on (Kamari and Ham 2020), the volumetric measurement framework can be summarized as follows: 1) resampling the point cloud model, 2) ground registration, and 3) computing the summation of distances between the resampled point cloud and the registered ground. Equation (2) demonstrates the volume of debris based on the aforementioned expressions.

$$V_{pwd} = GS^2 \times \sum_{i=1}^{n} \sum_{j=1}^{m} Z_{(X_i, Y_j)} \quad (2)$$

where $Z$ denotes the difference between resampled model and the registered ground at the location $(X_i, Y_j)$ and $GS$ is the resampling grid size. Upon identifying the type, location, and volume of debris, we assess the associated risk given the wind speed through kinetic energy equation (Wills et al. 2002), expressed as follows:

$$KE = \frac{1}{2}\, \rho_{pwd} V_{pwd}\, U^2 \tag{3}$$

where $KE$ indicates the kinetic energy of the projectile, and $U$ is the speed of the wind event. The parameter $\rho_{pwd}$ is the density of the debris and $V_{pwd}$ is the debris volume obtained by the volumetric measurement module. The higher level of kinetic energy implies a higher possibility of damage of projectiles in severe wind events. Given the severity of the wind event, at each location of the point cloud model, we solve for $KE$ to compute the risk map.

**EVALUATION AND EXPERIMENTAL SETUP**

A case study has been performed to demonstrate the fidelity of the proposed method and highlight its potentials. The jobsite presented in the case study accommodates loose and unprotected materials. A total of 186 aerial images have been collected using an Unmanned Aerial Vehicle (UAV). Examples of collected visual data, as well as a semantic map and depth map, are shown in Figure 3. The point cloud model of the at-risk jobsite presented in the case study is shown in Figure 4.a, and finally, the outcome of the depth-informed projection from all of the cameras onto the point cloud model is shown in Figure 4.b. The categories that account for the type of potential wind-borne debris are presented in the 3D semantic digital twin model. In this research, target semantic objects are loose materials or temporary facilities, such as metal girder, portable toilet, PVC piping, plywood, and metal piping. Categories except for those are considered as background objects during the semantic segmentation process.

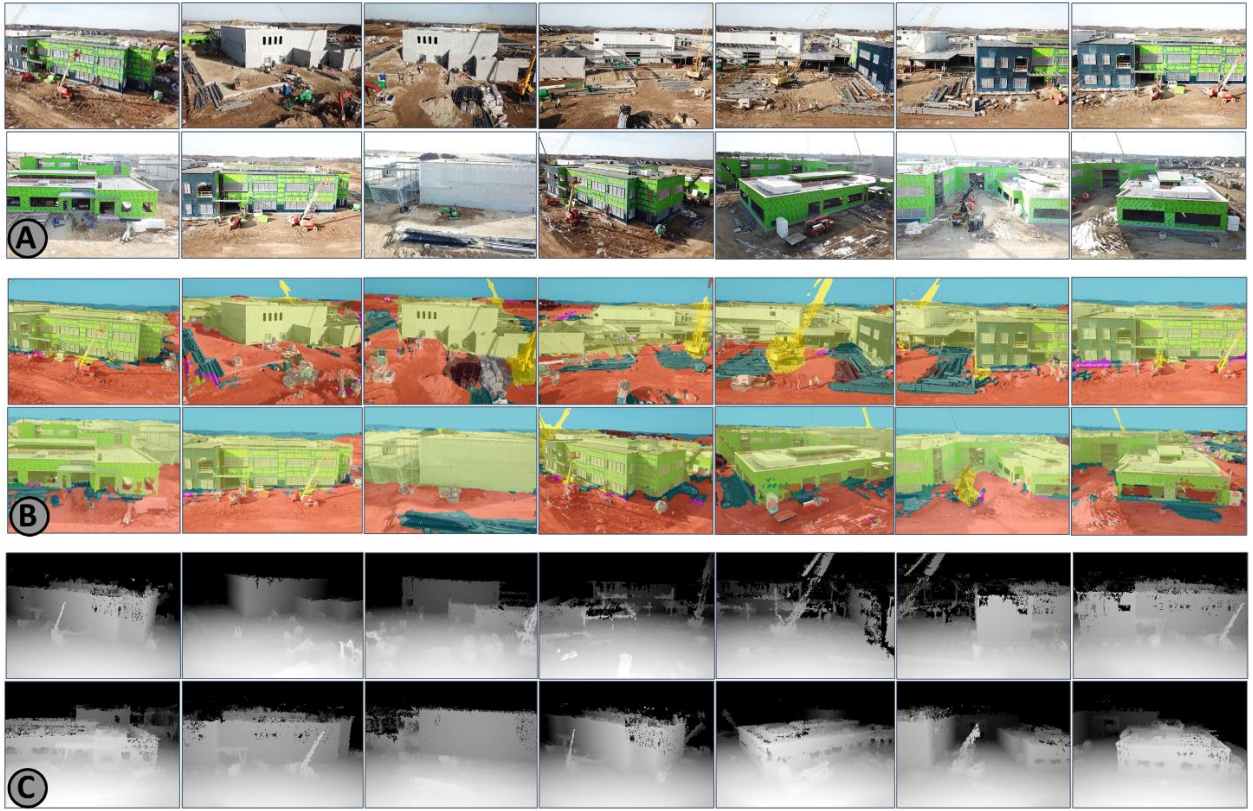

**Figure 3. a) collected aerial images, b) semantic map, and c) depth map.**

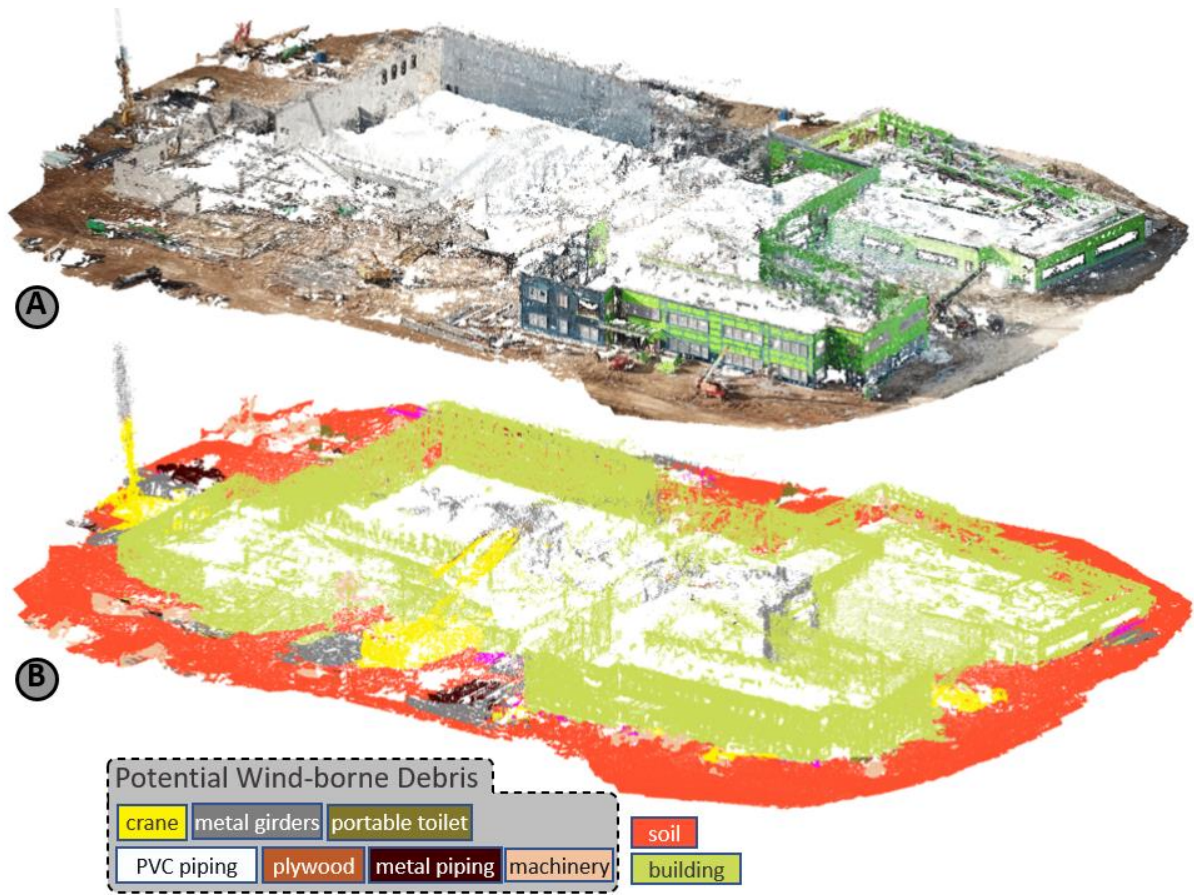

**Figure 4. Potential wind-borne debris through semantic 3D digital twin modeling.**

For point cloud generation, computation time with a mid-end commercial-level computing station, equipped with a single unite GTX 1080ti GPU, was about 14 minutes to reconstruct 831,682 number of points. Also, 2D semantic segmentation, and depth map generation among 186 images, along with depth-aware projection, were computed in 2, 9, 16 minutes, respectively. Finally, heatmaps are computed based on Equation 5. Figure 5 demonstrates the risk of potential wind-borne debris with respect to the intensity of the wind event. Wind speeds corresponding to the Saffir-Simpson hurricane wind scale (Taylor et al. 2010) are obtained, and the associated heatmaps are visualized for different hurricane categories.

As observed in Figure 5, certain debris do not pose a risk at lower wind speeds, but in higher wind speeds, such debris constitute a higher level of threat. Such insights obtained from the proposed systematic approach are not offered by hurricane preparedness checklists. This implies that the proposed method enables practitioners, safety directors, and contractors to effectively see through the potential threat and perform effective risk-based proactive measures for hurricane preparedness.

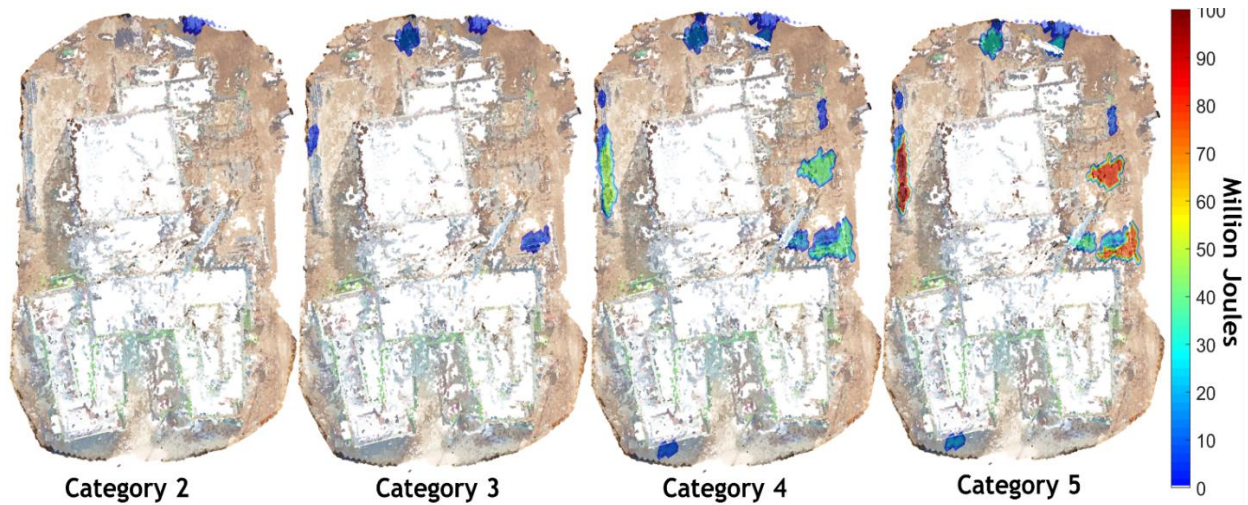

**Figure 5. Kinetic energy-based risk assessment with respect to different hurricane categories.**

## CONCLUSION

As hurricanes devastate the majority of southeastern states in the US, contractors and practitioners are required to remove, secure, and relocate the potential wind-borne debris in their jobsites. Failing to perform hurricane preparedness measures is classified as a second-degree misdemeanor in some states, and violators, including contractors, owners, and practitioners, are subjected to fines and imprisonment. Thus, given that there is a limited time to perform hurricane preparedness, a systematic way to identify the potential wind-borne debris based on spatial and temporal characteristics of jobsites is crucial for the construction industry. To address this demand, in this work, a novel vision-based method is proposed to locate and identify the type of potential wind-borne debris in construction jobsites. The proposed method is built upon two main modules of 1) semantic 3D digital twinning, and 2) risk assessment of the potential wind-borne debris. The outcome of the proposed method is a heatmap of the at-risk environment, which visualizes the unprotected materials and debris with respect to the severity of the wind event. The proposed technique paves the way to make risk-informed decisions and perform proactive measures in the context of hurricane preparedness. As future research, we are interested in validating our prototype in additional real-world examples and investigate how such quantitative risk could reduce the potential wind-induced damages in construction jobsites and neighboring communities.

## ACKNOWLEDGMENT

This material is based upon work supported by the National Science Foundation (NSF) under Grant No. 1832187. Any opinions, findings, and conclusions, or recommendations expressed in this material are those of the authors and do not necessarily reflect the views of the NSF.

## REFERENCES
Bohm, S. B. (2020). "Can Florida's Construction Sites be Hurricane-Proofed? Accessed Date May 2020."

CLM (2018). "A Perfect Storm: Hurricanes and Construction Defect Claims." Accessed Date: May 2021.
Farhadmanesh, M., Cross, C., Mashhadi, A. H., and Rashidi, A. (2021). "Highway Asset and Pavement Condition Management using Mobile Photogrammetry." *Transportation Research Record*.
FEMA (2012). "Miami International Airport-South Terminal." Accessed Date: May 2021.
Fermino (2013). "Sandy caused $185M in damage to WTC site."
Ghorbani, Z., and Behzadan, A. H. (2020). "Identification and Instance Segmentation of Oil Spills Using Deep Neural Networks." *CSEE 2020*.
Gregg, C. E., Houghton, B. F., Johnston, D. M., Paton, D., and Swanson, D. (2004). "The perception of volcanic risk in Kona communities from Mauna Loa and Hualālai volcanoes, Hawai."
Ham, Y., and Kamari, M. (2019). "Automated content-based filtering for enhanced vision-based documentation in construction toward exploiting big visual data from drones." *AutoCon*.
Kamari, M., and Gunes, O. (2016). "Segmentation and Analysis of a Sketched Truss Frame Using Morphological Image Processing Techniques."
Kamari, M., and Ham, Y. "Automated Filtering Big Visual Data from Drones for Enhanced Visual Analytics in Construction." *Proc., CRC 2018*.
Kamari, M., and Ham, Y. (2020) "Analyzing Potential Risk of Wind-induced Damage in Construction Sites and Neighboring Communities using Large-scale Visual Data from Drones." *Proc., CRC 2020*.
Kamari, M., and Ham, Y. (2020). "Vision-based volumetric measurements via deep learning-based point cloud segmentation for material management in jobsites." *Automation in Construction*.
Kim, J., and Ham, Y. "Vision-Based Analysis of Utility Poles Using Drones and Digital Twin Modeling in the Context of Power Distribution Infrastructure Systems." *Proc., CRC 2020*.
Kim, J., Kamari, M., Lee, S., and Ham, Y. (2021). "Large Scale Visual Data-Driven Probabilistic Risk Assessment of Utility Poles regarding the Vulnerability of Power Distribution Infrastructure System." *Journal of Construction Engineering and Management*.
Letchford, C., Norville, H., and Bilello, J. (2000). "Damage Survey and Assessment of Fort Worth"
Lowe, D. G. (2004). "Distinctive image features from scale-invariant keypoints." *International Journal of Computer Vision*, 60, 91-110.
Minor, J. E. (2005). "Lessons learned from failures of the building envelope in windstorms." *Journal of Architectural Engineering*.
OSHA (2019). *Occupational Health and Safety Administration*, Accessed Date: May 2021.
Sherafat, B., Ahn, C. R., and Akhavian, R. (2020). "Automated Methods for Activity Recognition of Construction Workers and Equipment: State-of-the-Art Review."
Sherafat, B., and Rashidi, A. (2019). "Automated activity recognition of construction equipment using a data fusion approach." *Computing in civil engineering*, 1-8.
Sherafat, B., and Rashidi, A. "A Software-Based Approach for Acoustical Modeling of Construction Job Sites with Multiple Operational Machines." *Proc., CRC 2020*, 886-895.
Taylor, H. T., Ward, B., Willis, M., and Zaleski, W. (2010). "The Saffir-Simpson hurricane wind scale." *Atmospheric Administration: Washington*.
Wills, J., Lee, B., and Wyatt, T. (2002). "A model of wind-borne debris damage." *Journal of Wind Engineering and Industrial Aerodynamics*, 90(4-5), 555-565.